\documentclass[11pt,a4paper]{article}
\usepackage[hyperref]{naaclhlt2018}
\usepackage{times}
\usepackage{latexsym}

\usepackage{color}
\usepackage{amsmath}
\usepackage{cases}
\usepackage{graphicx}
\usepackage{booktabs}
\usepackage{multicol}
\usepackage{xspace}

\usepackage{url}

\newcommand{\eg}{{\it e.g.}\xspace}
\newcommand{\ie}{{\it i.e.}\xspace}
\newcommand{\myquote}[1]{``#1''}

\newcommand{\attentivelm}{\emph{Attentive} RNN-LM\xspace}
\newcommand{\attentivelms}{\emph{Attentive} RNN-LMs\xspace}
\newcommand{\singscore}{{\emph{sin\-gle} score}\xspace}
\newcommand{\combscore}{{\emph{com\-bi\-ned} score}\xspace}

\aclfinalcopy 

\title{Persistence pays off: Paying Attention to\\What the LSTM Gating Mechanism Persists}
\author{
  Giancarlo D. Salton \and John D. Kelleher\\
  ADAPT Research Centre \\
  Dublin Institute of Technology \\
  Ireland \\
  {\tt giancarlo.salton@adaptcentre.ie john.d.kelleher@dit.ie}
 \\}

\date{}

\begin{document}
\maketitle
\begin{abstract}
  Language Models (LMs) are important components in several Natural Language Processing systems. Recurrent Neural Network LMs composed of LSTM units, especially those augmented with an external memory, have achieved state-of-the-art results. However, these models still struggle to process long sequences which are more likely to contain long-distance dependencies because of information fading and a bias towards more recent information. In this paper we demonstrate an effective mechanism for retrieving information in a memory augmented LSTM LM based on attending to information in memory in proportion to the number of timesteps the LSTM gating mechanism persisted the information.
\end{abstract}

\section{Introduction}
\label{sec:intro}
Language Models (LM) are important components in Natural Language Processing systems including, Statistical Machine Translation and Speech Recognition \cite{schwenk:2012}. An LM is generally used to compute the likelihood of a sequence of $N$ words appearing in a given language by using the chain rule

\begin{align}
p(w_1, \dots, w_N) = \prod_{t=1}^{N} p(w_n|w_1, \dots, w_{n-1})
\end{align}\label{eq:chain_rule}

Recently, Recurrent Neural Networks LMs (RNN-LMs) have became the state-of-the-art approach to language modelling \cite{jozefowicz:2016}. However, RNN-LMs struggle to keep their level of performance as the length of the input sequence increases, especially if the input contains long-distance dependencies (LDD).

A typical RNN-LM sequentially propagates forward a context vector that integrates information about previous inputs to use for the next prediction. Consequently, the information that is captured at the beginning of a sequence containing an LDD is likely to have faded from the context by the time the model spans that dependency. Another problem with such models is that the context vector may be dominated by more recent information. To address these limitations, several \myquote{memory-augmented} RNN-LMs architectures have been developed. In general, these models store the hidden states of the RNN in a memory buffer and then attempt to retrieve relevant information from the buffer at each timestep (e.g., \newcite{tran:2016}, \newcite{cheng:2016}, \newcite{daniluk:2017}, \newcite{merity:2017}, \newcite{grave:2017} and \newcite{salton:2017}).




In this paper, we analyse the behaviour of the LSTM units and demonstrate that an efficient and effective mechanism for a memory augmented LSTM based LM (LSTM-LM) to retrieve important information from its history is to construct a representation of the LSTM state history that weights information in proportion to the number of timesteps the LSTM persisted the information. Using this strategy reinforces the decisions of the LSTM gating mechanism at each timestep regarding what is important in a sequence. We demonstrate that using this simple strategy a memory augmented LSTM-LM can achieve state-of-the-art results for a single model on the Penn Treebank \cite{marcus:1994} with fewer parameters than its competitors, an also achieves near state-of-the-art results on the wikitext2 \cite{merity:2017}.


Structure: \S\ref{sec:lstm} reviews LSTMs and the gating mechanism; \S\ref{sec:uniform} discusses the effect of uniformly weighting the hidden states of an LSTM; \S\ref{sec:persistence} illustrates persistence of information in an LSTM; \S\ref{sec:average} describes our memory augmented LSTM-LM; \S\ref{sec:experiments} presents experiments and results; \S\ref{sec:discussion} contextualizes our findings; and \S\ref{sec:conclusions} contains conclusions.

\section{LSTMs}
\label{sec:lstm}
LSTM units (\emph{aka.} LSTM cells) and their variants (\eg, GRUs \cite{cho:2014}) are now a normal building block for neural based NLP systems \cite{bradbury:2016,murdoch:2017}. LSTMs retain and propagate information through the dynamics of the LSTM memory cell, hidden state, and gating mechanism (including the \emph{input}, \emph{forget}, and \emph{output} gates). The LSTM memory cell retains information that is only known by the unit itself and the hidden state shares information to other LSTM units in the same or any next layer of the network. This way, the units can decide what to keep in memory and how much of that information it wants the other units/layers to know about it. If something is deemed important, the units will both keep it in memory and let other units/layers to know about it. The gating mechanism controls the flow of information between the memory cell and the hidden state. Therefore, the gating mechanism plays an important role on the LSTM hidden dynamics.


The computations of a standard LSTM unit \cite{gers:2000} (without \textit{peephole connections}) involve iterating over the following equations

\begin{align}
  \mathbf{\widetilde{c}}_t = & tanh (\mathbf{W}\mathbf{x}_t + \mathbf{W}\mathbf{h}_{(t-1)} + \mathbf{b}) \label{eq:candidate}  \\
  \mathbf{i}_t = & \sigma (\mathbf{W}_{ii}\mathbf{x}_t + \mathbf{W}_{hi}\mathbf{h}_{(t-1)} + \mathbf{b}_i) \label{eq:input}  \\
  \mathbf{f}_t = & \sigma (\mathbf{W}_{if}\mathbf{x}_t + \mathbf{W}_{hf}\mathbf{h}_{(t-1)} + \mathbf{b}_f) \label{eq:forget} \\
  \mathbf{o}_t = & \sigma (\mathbf{W}_{io}\mathbf{x}_t + \mathbf{W}_{ho}\mathbf{h}_{(t-1)} + \mathbf{b}_o) \label{eq:output}
  \\
  \mathbf{c}_t = & \mathbf{f}_t \times \mathbf{c}_{(t-1)} + \mathbf{i}_t \times \mathbf{\widetilde{c}_t} \label{eq:block}
  \\
  \mathbf{h}_t = & \mathbf{o}_t \times \tanh (\mathbf{c}_t) \label{eq:state}
\end{align}

\noindent where the weight matrices $\mathbf{W}_{i*}$ are associated to the input; the weight matrices $\mathbf{W}_{h*}$ are associated with the recurrence; the vectors $\mathbf{i}_t$, $\mathbf{f}_t$, $\mathbf{o}_t$ are the activation vectors produced by the \textit{input}, \textit{forget} and \textit{output gates} respectively; $\mathbf{\widetilde{c}}_t$ is the candidate memory cell state; $\mathbf{c}_t$ is the new memory cell state; and $\mathbf{h}_t$ is the output of the unit. Equation \ref{eq:candidate} will produce a \textit{candidate vector} $\mathbf{\widetilde{c}}_t$ that contains information extracted from the input to the LSTM. The \textit{input gate} (Equation \ref{eq:input}) and \textit{forget gate} (Equation \ref{eq:forget}) compute an activation vector to be used to update the memory cell (Equation \ref{eq:block}): the \textit{candidate vector} is multiplied by the \textit{input gate} to decide how much of the input is important to the memory cell; and the content of the memory cell $\mathbf{c}_{(t-1)}$ is multiplied by the \textit{forget gate} to decide how much the memory cell will keep from its own content; the results of these two multiplications are merged to decide what will be remembered in the memory cell $\mathbf{c}_t$ for the next iteration.  The \textit{output gate} (Equation \ref{eq:output}) also calculates an activation vector that is multiplied by the current content on the memory cell $\mathbf{c}_t$ to produce what we call hidden state $\mathbf{h}_t$ (Equation \ref{eq:state}). This last step decides how much of the content in the memory cell $\mathbf{c}_t$ will be known on the next timestep (and by cells in the next layer if it is a multi-layered LSTM or to any layer that may come next o the network). It is important to notice that all the gates compute activations in the range $[0, 1]$ as they are using a \textit{sigmoid} activation function. If the values are close to 0, the gates are closed and if the values are close to 1, the gates are open. For example, if the \textit{forget gate} has a value close to 0 the content of the memory cell is erased; if the forget gate is close to 1, all the content of the memory cell will be exposed to other units or layers in the network.

The success of LSTM-RNNs is attributed to their ability to retain information about the input sequence for several timesteps in their internal memory cell $\mathbf{c}_t$. That information is then made available to the next layer in the network for the amount of timesteps it is considered relevant to the current sequence. As pointed by \newcite{murdoch:2017}, each input to an LSTM makes a contribution to the hidden state of the LSTM and that is reflected when Equation \ref{eq:block} is iterated. At any given timestep $t$, the cell state $\mathbf{c}_t$ can be decomposed into

\begin{align}
 \mathbf{c}_t = \sum_{i=1}^{t} (\prod_{j=i+1}^{t}\mathbf{f}_i)\mathbf{i}_i\mathbf{\widetilde{c}}_i
\end{align}

\noindent which, according to the authors, can be interpreted as the contribution at timestep $t$ to the memory block $\mathbf{c}_t$ by a particular past input at timestep $j$. In that view, the contribution of an input to a given timestep can be understood as an importance score weighted by the LSTM's gating mechanism. Therefore, if something is important to the current context if should receive a larger importance score and be held in the memory block for a number timesteps. In addition to retaining information, \newcite{murdoch:2017} have also demonstrated that, despite the fact that it is still difficult to interpret what specific activations in the hidden dynamics of LSTM units mean, it is possible to extract semantically meaningful rules from the memory cells to train a powerful classifier that can approximate the output of the LSTM itself. Moreover, \newcite{strobelt:2016} and \newcite{karpathy:2016} have demonstrated that these networks can extract meaningful attributes from the data into the memory cells. These attributes carry fine grained information and keep track of attributes such as line lengths, quotes and brackets.

Although these and other work demonstrate the power of LSTM units and their gating mechanism, RNN-LMs based on such units (LSTM-LMs) struggle to process long sequences. In our view, the main reason for this degradation in performance happens exactly because of the hidden state dynamics of the LSTM units. Once the information retained in the memory cell $\mathbf{c}_t$ is outdated, the \textit{forget gate} $\mathbf{f}_t$ erases that block enabling the unit to store fresh data without interference from previous timesteps \cite{gers:2000,gers:2003}. This behaviour creates a natural bias towards more recent inputs given that the memory cell has limited capacity to store previous information and, once the memory cell is saturated, the \textit{forget gate} will start to drop information in favour of more recent inputs. Even though the LSTM units can learn which information it must retain and for how long, the model will struggle with long sequences that are more likely to contain LDDs and that saturate the memory cell. 

Once a memory cell has been saturated then, although some content has received a large importance score in past steps, it may be dropped from the memory cell (because of the inherent limitation of the LSTM's capacity of storing content) and will not be available to contribute to the next steps. For example, an LSTM-LM trained on English may persist the information related to a subject of a sentence sentence for a number of time steps because the subject is important but this information may still have faded by the time the verb is reached. However, by augmenting the network with a memory buffer the information relating to the subject continues to be accessible so long as the memory buffer is not reset. This behaviour is an indication of why the memory augmented models such as the Neural cache model of \newcite{grave:2017} and the Pointer LSTM of \newcite{merity:2017} has gained success and achieved state-of-the-art in LM research. Even though the required content has already faded from the context, the memory augmentation make it available for subsequent timesteps.

\section{The Curious Effectiveness of Uniform Attention}
\label{sec:uniform}
As was noted in Section \ref{sec:intro}, in recent years a number of extensions to RNN-LMs have be proposed to overcome this fading of information from memory by adding a memory buffer (that is used to store the LSTM hidden state at each timestep) and then at each timestep construct a representation of this history to inform the current prediction. A variety of relatively sophisticated mechanisms for retrieving information from the memory buffer have been proposed. In many cases these retrieval mechanisms include an extra neural network in the RNN-LMs that at each timestep predicts what elements in the memory buffer should be retrieved.

\newcite{salton:2017} is a recent example that uses an extra  neural network\footnote{Similar to that proposed by \newcite{bahdanau:2015} and \newcite{luong:2015} for Neural Machine Translation (NMT)} to learn what to retrieve from memory. In this architecture at the end of a timestep the current LSTM hidden state is added to the memory buffer. Then at the beginning of the next timestep the additional neural network predicts an attention distribution over the elements of the memory buffer (\ie, the previous LSTM hidden states). Using this attention distribution a compact representation of the RNN-LMs history is constructed by calculating a weighted sum of the elements in the memory (where the weight of each element is the attention attributed to it by the RNN). Curiously, although this architecture was successful in terms of performance the attention mechanism did not work as expected. Instead of focusing attention for each time step on particular relevant elements in memory it spread out the attention nearly uniformly across the memory. It might appear that this architecture was using a strategy of \myquote{pay equal attention to everything in the past}. However, we argue this interpretation ignores the power of the LSTM gating mechanism.

Our interpretation of the uniform attention mechanism presented by \newcite{salton:2017} is that their \attentivelm is in fact (indirectly) reinforcing the decisions of the gating mechanism of the LSTM units and is retrieving information that is persisted across multiple timesteps. This is important because it indicates that it may be more fruitful and efficient to leverage the decisions made by the LSTM gating mechanism (decisions that the network must make anyway) to drive the retrieval of information from the memory buffer rather than train a separate neural network. It is worth emphasising that to date none of the different retrieval mechanisms proposed in the literature on memory augmented LSTM-LMs have explicitly considered the behaviour of the LSTM gating mechanism.

\section{The Persistence of Information}
\label{sec:persistence}
\begin{figure}[t]
\includegraphics[width=0.75\linewidth]{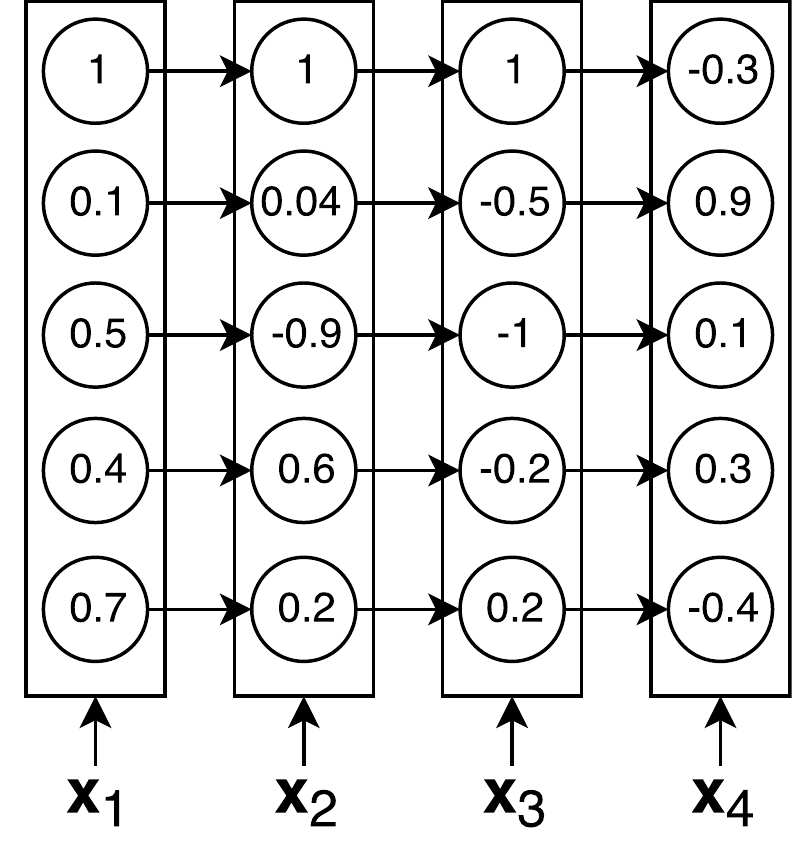}
\centering
\caption{Illustration of the persistence of information in an LSTM-RNN. At any given timestep the \textit{forget gate} may decide to erase the information stored in the memory cell of the LSTM unit and allow the \textit{input gate} to decide what to store in its place. If something is deemed important by the unit, it is held for more than one timestep until it is regarded as outdated. In addition, if the information is important, the \textit{output gate} will let it flow to the next timesteps and other units in the network. In this hypothetical example, considering the \textit{output gate} is open (\ie, values of 1), the information of the top memory cell is maintained for timesteps 1, 2 and 3, whilst the information on the bottom memory cell is held on timesteps 2 and 3. All the other information is held only for a single timestep. Also, notice that we drop the initial state $\mathbf{h}_0$ from this image}
\label{fig:presistence}
\end{figure}

The LSTM gating mechanism will attempt to persist important information for as long as possible (or until the state is saturated). Figure \ref{fig:presistence} illustrates the idea of persistence within an LSTM unit and how information is eventually dropped. In this illustration, we can see that the information contained in the top of the LSTM block is persisted across 3 timesteps (timesteps 1, 2, and 3) and the information at the bottom LSTM block is persisted for 2 timesteps (timesteps 2 and 3). All other content was held for a single timestep. However, at the moment the model is making a prediction on the last timestep (in this case, timestep 4), all the old information (including the information that was persisted) has already vanished from the context.  


We propose that when retrieving/constructing a representation of the LSTM history from a memory buffer the information held for more than one timestep should be weighted in proportion to the number of timesteps the LSTM gating mechanism persisted it across. In the specific situation illustrated in Figure \ref{fig:presistence} we argue that the information persisted in the top memory cell for 3 timesteps should have a larger during weighting in the representation of the sequence history compared to the pieces of information that were only held for a single timestep.  This way, we let the gating mechanism of the LSTM determine what is important about the input and, anything that is persisted for more than one timestep, will have a greater impact on the final prediction even if that information has already faded from the current context.

A simple and efficient way to implement this strategy is at each time point to construct a representation of the history of the RNN-LM that is simply an average of the LSTM hidden states in the memory buffer. Pieces of information that the LSTM unit persists for several time steps will have a bigger impact on this average (simply because they are included multiple times) relative to items that are not persisted. In effect, this average weights each piece of information in proportion to the number of time steps the LSTM persisted it and so an RNN-LM that uses this average as its representation of history pays attention to what the LSTM gating mechanism persisted.


\section{Averaging the Outputs}
\label{sec:average}
In this work we adapt and simplify the architecture of \newcite{salton:2017} and use a simple average of previous outputs instead of a neural network based attention mechanism. Our intuition for this modification is that the gating mechanism of the LSTM is telling us what is important about an input and that we must find a way to make that information available for long distances in the future. In fact, \newcite{ostmeyer:2017} have presented a model that computes a recurrent weighted average (RWA) over every past hidden state. However, the authors limit themselves to evaluate the model over simple tasks and the effectiveness of that model over language modelling is still to be demonstrated.


Compared to other memory augmented models our architecture is relatively simple. Figure 2 displays a timestep for our system\footnote{Please note that this figure is inspired by the architecture schematic in \newcite{salton:2017}}. A multi-layered LSTM-RNN encodes an input at each timestep and the outputs of the last recurrent layer (\ie, its hidden state) $h_t$ (Equation \ref{eq:state}) is added to memory. At each timestep an average of the vectors in the memory buffer is calculated and concatenated with the $h_t$ generated by the processing of the current input. This concatenated vector is then feed into the softmax layer which predicts the distribution for the next word in the sequence.

\begin{figure}[t]
\includegraphics[width=\linewidth]{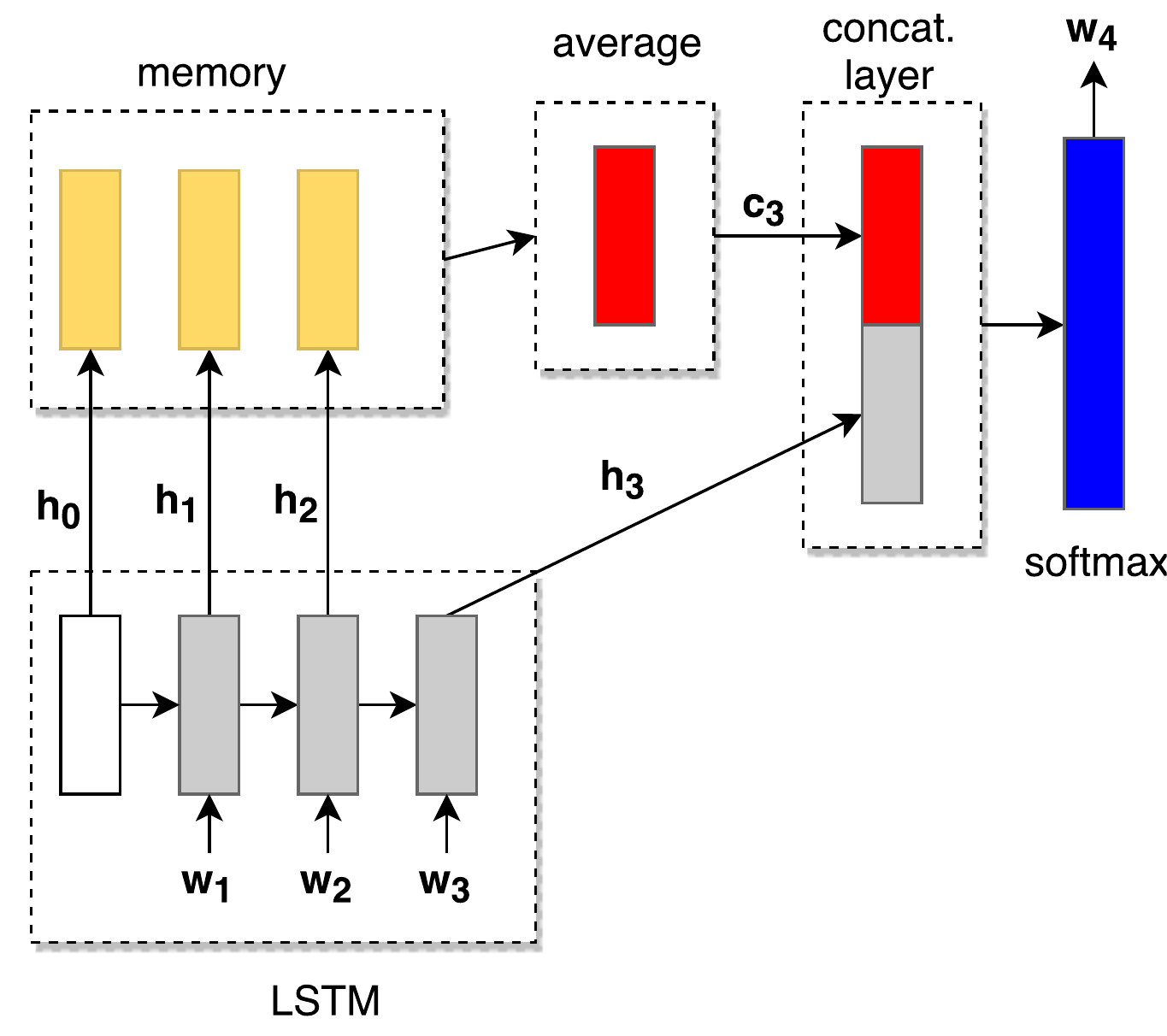}
\centering
\caption{Illustration of a step of the Average LSTMLM. In this example, the model receives the third word as input ($\textbf{w}_{3}$) after storing the previous states ($\textbf{h}_{1}$, $\textbf{h}_{2}$ and the initial state $\textbf{h}_{0}$) in memory. After producing $\textbf{h}_{3}$, the model computes the context vector (in this case $\textbf{c}_{3}$) by averaging the states stored in memory. $\textbf{c}_{3}$ is concatenated to $\textbf{h}_{3}$ before the softmax layer for the prediction of the fourth word $\textbf{w}_{4}$. Also note that $\mathbf{h}_3$ is stored in memory only at the end of this process and does not participate in the $\textbf{c}_{3}$ calculation.}
\label{fig:model}
\end{figure}

More formally, the hidden state at timestep $t$, $\mathbf{h}_t$, is calculated for an input $\mathbf{x}_t$ as

\begin{align}
\mathbf{h}_t = g(\mathbf{x}_t, \mathbf{h}_{t-1})
\label{eq:hidden}
\end{align}

\noindent where $g$ is a LSTM-RNN. Instead of applying a sophisticated attention mechanism to the hidden states ($\mathbf{h}_{<t}$), we calculate the context vector $\mathbf{c}_t$ as the average of the hidden states in memory:

\begin{align}
\mathbf{c}_t = \frac{1}{t}\sum_{i=0}^{t-1} \mathbf{h}_i
\label{eq:context}
\end{align}

\noindent where each $\mathbf{h}_i$ is a hidden state stored in the memory buffer\footnote{Please notice that the iteration index for the summation starts at 0 and, thus, the length of the memory buffer is equal to the timestep index $t$. As we shall explain later, we also included the initial state $\mathbf{h}_0$ in the memory buffer and count it to calculate the average.}. After this step, $\mathbf{c}_t$ is concatenated to the current hidden state $\mathbf{h}_t$ by means of a concatenation layer

\begin{align}
\mathbf{h}_{t}^{\prime} = tanh(\mathbf{W}_c[\mathbf{h}_t;\mathbf{c}_t] + \mathbf{b}_t)
\label{eq:concat}
\end{align}

\noindent and, finally, the modified hidden state $\mathbf{h}_{t}^{\prime}$ is forwarded to the softmax layer for the next prediction

\begin{align}
p(w_t|w_{<t}, x) = softmax(\mathbf{W}_s\mathbf{h}_{t}^{\prime}  + \mathbf{b})
\label{eq:probability}
\end{align}

In our experiments with this uniform attention, we found that initialising the memory with a zero vector $\mathbf{h}_{0}$ and allowing the model to count this vector as part of the memory when calculating the average\footnote{In other words, the index of the memory starts at timestep 0 instead of timestep 1. Thus, the memory at any given timestep $t$ will be of length $t+1$.} improved the performance of the model.


\section{Experiments}
\label{sec:experiments}
To test our intuitions, we evaluate the averaging process of the model using the PTB dataset  using the standard split and pre-processing as in \newcite{mikolov:2010} which consists of 887K, 70K and 78K tokens on the training, validation and test sets respectively. We also evaluate the model on the wikitext2 dataset using the standard train, validation and test splits which consists of around 2M, 217K tokens and 245k tokens respectively.

\subsection{PTB Setup}
\label{sub:wiki_setup}

Following \newcite{salton:2017} we trained a multilayer LSTM-RNN with 2 layers of 650 units for the PTB experiment. We trained them using Stochastic Gradient Descent (SGD) with an initial learning rate of 1.0 and we halved the learning rate at each epoch after 12 epochs. We train the model to minimise the average negative log probability of the target words until we do not get any perplexity improvements over the validation set with an early stop counter of 10 epochs. We initialize the weight matrices of the network uniformly in [$-0.05$, $0.05$] while all biases are initialized to a constant value at $0.0$ with the exception of the \textit{forget gate} biases which is initialised at $1.0$ as suggested by \newcite{jozefowicz:2015}. We also apply $50\%$ dropout \cite{srivastava:2014} to the non-recurrent connections and clip the norm of the gradients, normalized by the mini-batch size of $32$, at $5.0$. We also tie the weight matrix $\mathbf{W}_s$ in Equation \ref{eq:probability} to be the embedding matrix as in \newcite{press-wolf:2016}. Thus, the dimensionality of the embeddings is  set to 650.

\subsection{wikitext2 Setup}
\label{sub:wiki_setup}

For the wikitext2 experiments we trained a multilayer LSTM-RNN with 2 layers of 1000 units. We also used SGD to minimise the average negative log probability of the target words with an initial learning rate of 1.0. We decayed the the learning rate by a factor of 1.15 at each epoch after 14 epochs and we used an early stop counter of 10 epochs. Similarly to the PTB experiment, we initialize the weight matrices of the network uniformly in [$-0.05$, $0.05$] while all biases are initialized to a constant value at $0.0$ with the exception of the \textit{forget gate} biases which is initialised at $1.0$. For this model we apply $65\%$ dropout to the non-recurrent connections and clip the norm of the gradients, normalized by the mini-batch size of $32$, at $5.0$. Once again, we tie the weight matrix $\mathbf{W}_s$ in Equation \ref{eq:probability} to be the embedding matrix. Thus, the dimensionality of the embeddings is set to 1,000.

\subsection{Data Manipulation and Batch Processing}
\label{sub:batch}

When training each model, we use all sentences in the respective training set, but we truncate all sentences longer than 35 words and pad all sentences shorter than 35 words with a special symbol so all have the same length. We use a vocabulary size of 10k for the PTB and 33,278 for the wikitext2. Each of the mini-batches we use for training are then composed of 32 of these sentences taken from the dataset in sequence. 

Contrary to the recent trend in the field, we do not allow successive mini-batches to sequentially traverse the dataset. We reinitialize the hidden state of the LSTM-RNN at the beginning of each mini-batch, by setting it to all zeros.  Our motivation for not sequentially traversing the dataset is that although sequentially traversing has the advantage of allowing the batches to be processed more efficiently, some dependencies between words may not be learned if batch traversing is in use as the mini-batch boundaries can split sentences. We also found that allowing the initial state of all zeros to be included in the memory when averaging improves the performance of the Average RNN-LM.



\subsection{Results}
\label{sec:results}

\begin{table*}[t]
  \begin{tabular*}{\textwidth}{lccc}
    \toprule
      \textbf{Model} & \textbf{Params} & \textbf{Valid. Set} & \textbf{Test Set} \\
    \midrule
    \textbf{Single Models} \\
    \midrule
      Large Regularized LSTM \cite{zaremba:2015}      & 66M  & 82.2  & 78.4             \\
      Large + BD + WT \cite{press-wolf:2016}          & 51M  & 75.8  & 73.2             \\
      Neural cache model (size = 500) \cite{grave:2017}  & -    & -  & 72.1     \\
      Medium Pointer Sentinel-LSTM \cite{merity:2017} & 21M  & 72.4  & 70.9              \\
      Attentive LM w/ \combscore function \cite{salton:2017}   & 14.5M  & 72.6  & 70.7              \\
      Attentive LM w/ \singscore function \cite{salton:2017}  & 14.5M  & 71.7  & 70.1     \\
      Averaging RNN-LM & 14.1M & 71.6 & \textbf{69.9} \\
      \midrule
      \textbf{Model Averaging} \\
      \midrule
        2 Medium regularized LSTMs \cite{zaremba:2015}  & 40M   & 80.6   & 77.0            \\
        5 Medium regularized LSTMs \cite{zaremba:2015}  & 100M  & 76.7   & 73.3             \\
        10 Medium regularized LSTMs \cite{zaremba:2015} & 200M  & 75.2   & 72.0             \\
        2 Large regularized LSTMs  \cite{zaremba:2015}  & 122M  & 76.9   & 73.6             \\
        10 Large regularized LSTMs \cite{zaremba:2015}  & 660M  & 72.8   & 69.5             \\
        38 Large regularized LSTMs \cite{zaremba:2015}  & 2508M & 71.9   & \textbf{68.7}    \\
  \end{tabular*}
  \caption{Perplexity results over the PTB. Symbols: WT = weight tying \cite{press-wolf:2016}; BD = Bayesian Dropout \cite{gal:2015}. Please note that we could not calculate the number of parameters for some models given missing information in the original publications.}
  \label{table:ptb_results}
\end{table*}

\begin{table*}[t]
  \begin{tabular*}{\textwidth}{lccc}
    \toprule
      \textbf{Model} & \textbf{Params} & \textbf{Valid. Set} & \textbf{Test Set} \\
    \midrule
      Zoneout + Variational LSTM \cite{merity:2017}       & 20M  & 108.7  & 100.9    \\
      LSTM-LM \cite{grave:2017}                           & -    & -      & 99.3     \\
      Variational LSTM \cite{merity:2017}                 & 20M  & 101.7  & 96.3     \\
      Neural cache model (size = 100) \cite{grave:2017}   & -    & -      & 81.6     \\
      Pointer LSTM (window = 100) \cite{merity:2017}      & 21M  & 84.8   & 80.8    \\
      Averaging RNN-LM & 50M & 74.6 & 71.3 \\
      Attentive LM w/ \combscore function  \cite{salton:2017}  & 51M  & 74.3   & 70.8     \\
      Attentive LM w/ \singscore function  \cite{salton:2017}  & 51M  & 73.7   & 69.7     \\
      Neural cache model (size = 2000) \cite{grave:2017}  & -    & -      & \textbf{68.9}     \\
  \end{tabular*}
  \caption{Perplexity results over the wikitext2. Please note that we could not calculate the number of parameters for some models given missing information in the original publications.}
  \label{table:wiki_results}
\end{table*}

Table \ref{table:ptb_results} presents the results in terms of perplexity of the models trained over the PTB dataset. As we can see, the results obtained by the Averaging RNN-LM are similar to those obtained by the \attentivelms of \newcite{salton:2017} and by an ensemble of 38 LSTM-LMs. Despite the simple method to retrieve information from the previous timesteps, the Averaging RNN-LM achieves the same level of performance of more complex models with less computation overhead.

Table \ref{table:wiki_results} presents the results in terms of perplexity of the models trained over the wikitext2 dataset. Although the Averaging RNN-LM is still behind the \attentivelms and the Neural cache model of \newcite{grave:2017} on this dataset, the results are encouraging given the simplicity of the Averaging RNN-LM.

\section{Discussion}
\label{sec:discussion}
The Averaging LSTM-LM achieves the lowest perplexity for a single model on the PTB (see Table \ref{table:ptb_results}). It does this using less parameters than any of the other models tested on the PTB\footnote{We have considered using the concept of a parameter budget \cite{collins:2016,melis:2017} to contextualize the difference in parameters but given that the difference in terms of total number parameters between the baseline \attentivelms and our model is relatively small (essentially arising from the fact that we have dropped the attention neural network component from that architecture) we felt it would not be appropriate to apply this concept.}. Given the similarity of the results between the \attentivelms of \newcite{salton:2017} and the Averaging LSTM-LM it would appear that our hypothesis that the \attentivelms was (indirectly) learning to use the dynamics of the LSTM gating mechanism is correct. The Averaging LSTM-LM is also very competitive on the PTB compared to the ensemble methods and in this case the difference in total number of parameters between the Averaging LSTM-LM and these ensemble models is significant. 


Focusing on the results for the wikitext2 dataset  (see Table 2) the Neural cache model is still the best performing model on this dataset. We are not able to estimate the number of parameters for the Neural cache model so we have not included the parameter size of that model in the table. In discussing the wikietext2 results it is worth noting that the \attentivelms of \newcite{salton:2017} and the Averaging LSTM-LM are the only models in Table \ref{table:wiki_results} that reset their memory at each sentence boundary whereas the memory buffers of other models were allowed to span sentence boundaries. This is an important difference for the wikitext2 dataset because it is composed off documents where the sentences are in the correct order. Consequently, by allowing the memory buffer to span sentence boundaries on the wikitext2 dataset these models were able to carry forward contextual information from preceding sentences and as a result they did not suffer from as severe a cold-start problem at the beginning of each sentence. Given this difference the competitive performance of the Averaging LSTM-LM is encouraging.

The results for the wikitext2 dataset highlights an interesting trade-off and design choice for memory augmented LSTM-LMs. One approach is to use a dynamic length memory buffer which resets at sentence boundaries and uses a simple mechanism, such as averaging, to construct a representation of the memory to inform the prediction at each timestep. This is the approach we have proposed in this paper. This approach has the advantages of simplicity (model size) and that the memory length can be anchored to landmarks in the history, such as sentence boundaries. This approach is most appropriate for sentence based NLP tasks such as sentence based Machine Translation. There is a question, however, regarding whether this approach will scale to very long sequences (such as documents) as averaging over long-histories may result in all histories appearing similar. We have done some initial experiments where we have permitted the memory buffer to hold longer sequences before being reset and the performance of the Averaging LSTM-LM dipped.  The alternative approach is to use a larger memory buffer and a more sophisticated retrieval mechanism, for example the Neural cache model of \cite{grave:2017} is a useful exemplar for this approach with a memory buffer of 2,000 steps. As the wikitext2 results demonstrate this second approach works well for large datasets where the sentences are in sequence, the cost of this approach being a more complex architecture.

\section{Conclusions}
\label{sec:conclusions}
In this paper we have highlighted the power of the LSTM gating mechanism and argued that the persistence dynamics of this mechanism can provide useful clues regarding what information is important within a sequence for language modelling. Previous works have demonstrated that LSTMs are capable of extracting important features about an input sequence and that the dynamics of the memory block calculates an importance factor for the contribution of each previous inputs for a given timestep. However, the dynamics of the gates may drop the information kept in memory in favour of more recent information. Once dropped information that may have been held for several timesteps vanishes from the context and is not available anymore to the model. To deal with this problem, a variety of \myquote{memory augmented} LSTM-LMs have been proposed which store the previous hidden states of an LSTM in order to make the vanished information available again to subsequent timesteps. However, to date none of these memory augmented LSTM-LMs have explicitly used the persistence of information within an LSTM unit to inform the decision regarding what information should be retrieved from the memory buffer. We argue that ignoring the internal dynamics of the LSTM is to overlook a useful and computationally cheap source of information for memory augmented language modeling.

We believe that attending to the information that an LSTM gating mechanism has decided is important in an input sequence at a given timestep (and hence has persisted to a later timestep) is a natural way of deciding what information will be useful again at a subsequent timestep. Even if the information contained in the LSTM is replaced or altered later in the process, we argue that it is relevant to the entire history in proportion to the amount of timesteps it was held. Informed by this hypothesis, in our work we demonstrated that a simple average of the previous LSTM hidden states in memory is an effective mechanism for providing information to the current timestep about previous inputs. 

Admittedly, rating the importance of information in terms of the number of timesteps the LSTM persisted it for is a relatively simplistic view of the dynamics of LSTM units and of the complexity of language. Furthermore, implementing this strategy using an average of past states is also a relatively blunt way of instantiating this approach. However, as our results demonstrate this simple approach is effective and we understand this is a starting point. By drawing attention to the signals implicit in the dynamics of LSTM units we hope to contribute to the development of more efficient LMs. At the same time, the fact that the internal dynamics of an LSTM unit may be used to explicitly signal what is important and what should be retrieved from a memory buffer may suggest alternative constraints and opportunities that should be considered in the design of neural units and by doing so contribute to the development of a new class of units for use in RNN-LMs.


\section*{Acknowledgments}


\bibliography{naaclhlt2018}
\bibliographystyle{acl_natbib}

\end{document}